\documentclass[letterpaper,10pt,conference]{ieeeconf}

\IEEEoverridecommandlockouts                              

\usepackage{graphics} % for pdf, bitmapped graphics files
\usepackage{epsfig} % for postscript graphics files
\usepackage{amsmath} % assumes amsmath package installed
\usepackage{amssymb}  % assumes amsmath package installed
\usepackage{bm}
\usepackage{algorithm}
\usepackage{algpseudocode}
\usepackage{float}
\usepackage{subfigure}
\usepackage{cite}
\usepackage{graphicx}
\usepackage{xcolor}
\usepackage{mathrsfs}
\usepackage{multirow}
\usepackage{booktabs}

\title{\LARGE \bf
Dynamics-Informed Reinforcement Learning for Agile and Energy-Efficient Locomotion of a Monopedal Hopping Quadcopter
}
\author{Ruigang Chen$^{1,2}$, Qi Zhang$^{1}$, Zhicheng Zhong$^{1}$, Zhuorui Yun$^{1}$, Yizhar Or$^{2}$, and Mingyi Liu$^{1,2}$
\thanks{*This work was supported by the startup fund from Guangdong Technion – Israel Institute of Technology (Corresponding author: Mingyi Liu).}%
\thanks{$^{1}$Department of Mechanical Engineering and Robotics, Guangdong Technion - Israel Institute of Technology, Shantou, 515063, Guangdong, China.}%
\thanks{$^{2}$Department of Mechanical Engineering, Technion-Israel Institute of Technology, Haifa, 3200003, Israel.}%
}

\begin{document}

\maketitle
\thispagestyle{empty}
\pagestyle{empty}

\begin{abstract}
Although aerial-legged robots offer combined agility and efficiency, controlling high-speed hopping under complex hybrid dynamics is challenging. Reinforcement Learning (RL) is promising but prone to energy-inefficient ``reward hacking". We propose a Dynamics-Informed RL framework for a monopedal hopping quadcopter. By embedding a target Specific Energy into the reward, we constrain the optimization to a physically viable energy manifold, ensuring stable hopping behaviour. By rewarding the phase-consistent behavior, it can encourage bio-inspired stance-phase impulse. Furthermore, penalizing the electro-mechanical power waste induces the motors generate an efficient impulse. This enables the policy to inject energy strictly during spring restitution without heuristic state machines. MuJoCo simulations validate robust height regulation and forward velocity tracking up to 2.0 m/s despite severe attitude-contact coupling. Ultimately, our approach yields a highly agile hopping gait, reducing energy consumption by 82\% and 73\% compared to hovering baselines and inefficiency baseline, respectively.
\end{abstract}

\section{Introduction}

Jumping, recognized as a highly efficient mode of locomotion, enables arboreal and semi-aquatic animals to navigate through dense vegetation, evade predation threats, and access essential resources in challenging ecosystems such as woodlands, tropical rainforests, and wetland habitats \cite{richards2017kinematic, gvirsman2016dynamics, brunt2016amphibious, jung2017effect}. Compared with walk and crawl, jump has advantages of high energy density, efficient obstacle negotiation, rapid
terrain transition and so on \cite{siwanowicz2017three}. This pursuit of versatile multi-modal locomotion has spurred significant interest in jumping robotics \cite{zhang2017survey}. By combining the obstacle negotiation of jumping with the energy efficiency of elastic elements, these hybrid systems offer robust solutions for unstructured environments \cite{zaitsev2015locust, haldane2017repetitive, csomay2023nonlinear}. Among these, the monopedal hopping quadcopter represents a minimalist yet highly challenging architecture \cite{zhu2022pogodrone, bai2024agile, li2025high}. It extends the classical Spring-Loaded Inverted Pendulum (SLIP) model \cite{saranli2010approximate, poulakakis2009spring} by introducing active aerial thrust, enabling extended operational endurance and aggressive maneuvers typically too risky for pure hopping platforms.

\begin{figure}[t]
    \centering
    
    \includegraphics[width=1\linewidth]{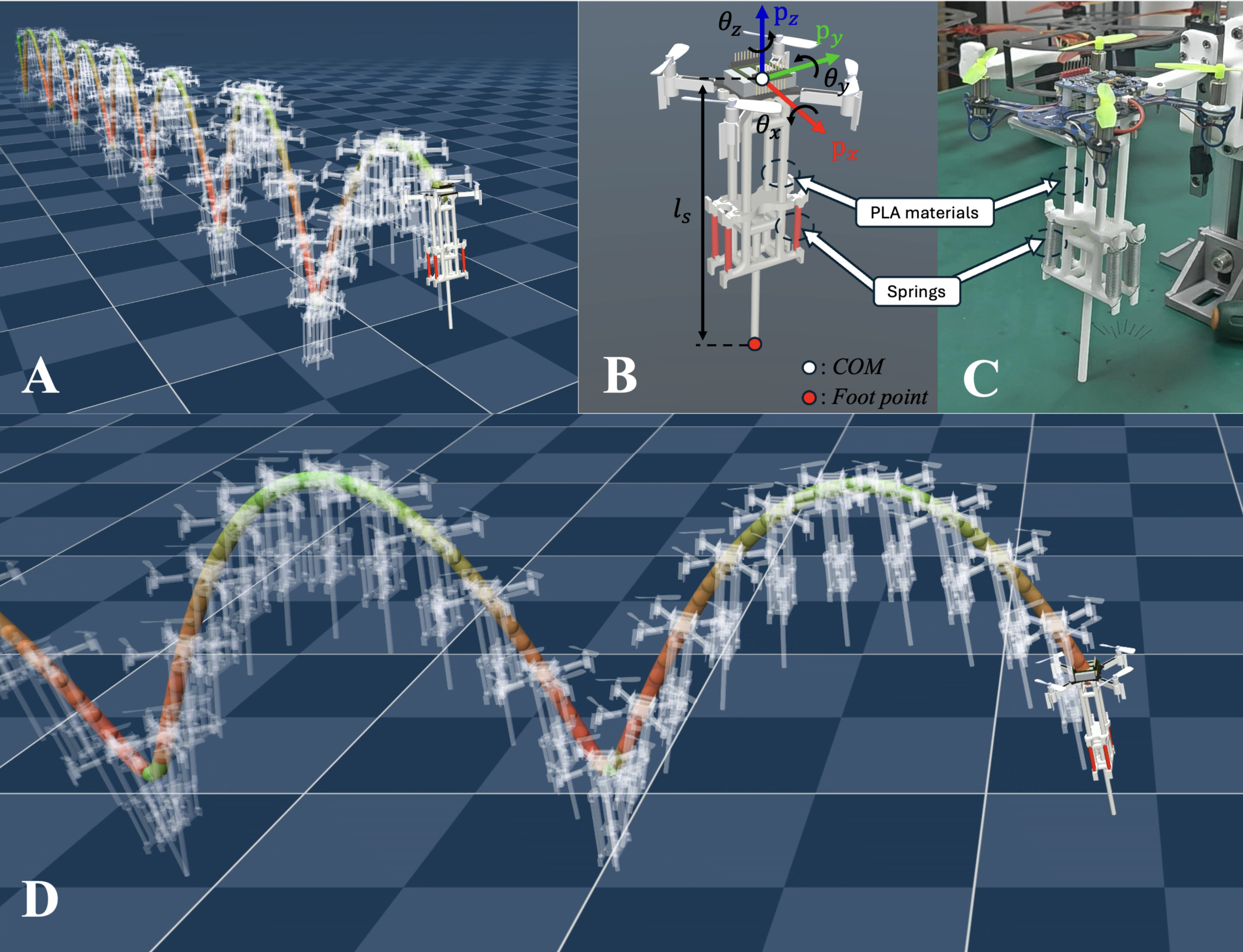} 
    \caption{\textbf{Visualized jump trajectory, mechanical structure, and coordinate definition} 
  (A) Chronophotography of the learned high-speed hopping gait ($v_x = 1.5$\,m/s). 
  (B) The proposed monopedal robot platform based on the Crazyflie 2.1 and its modeling description. 
  (C) Photograph of the prototype. Annotations explicitly indicate the four parallel springs responsible for storing and releasing impact energy, and the custom compliant leg base fabricated from 3D-printed Polylactic Acid materials. 
  (D) The visualization encodes velocity information: the hue of the robot's color trace is proportional to its instantaneous speed. Darker regions correspond to higher velocities, while lighter regions indicate lower velocities near the apex, which validates the effective energy exchange.}
  \label{fig:teaser}
\end{figure}

However, controlling this under-actuated, hybrid dynamical system \cite{goebel2009hybrid, raibert1986legged} is non-trivial. Unlike fully actuated legged robots \cite{he2020mechanism}, control forces are coupled entirely through the quadcopter's attitude \cite{bouabdallah2007design}, which becomes highly nonlinear during high-speed horizontal locomotion. Existing control strategies face significant bottlenecks. While recent model-based advancements attempt to capture full stance-phase nonlinearities using neural network-compressed models \cite{li2025high}, they still fundamentally rely on heuristic state machines and rigid phase transition logic \cite{bai2024agile, li2025high, huang2024real}. This discrete decoupling restricts the continuous operational envelope and struggles to orchestrate the severe attitude-contact coupling seamlessly, often causing divergence during continuous agile maneuvers.

Conversely, while RL \cite{arulkumaran2017deep, schulman2017proximal, muzio2022deep, kumagai2025reinforcement} circumvents explicit linearization, it struggles with behavioral alignment. Without physical constraints, RL agents routinely exploit reward loopholes \cite{yuan2019novel, knox2024learning, ibrahim2024comprehensive} to ``hover-hop''—continuously thrusting mid-air to correct kinematic errors rather than exploiting natural ballistic coasting. This contrasts starkly with biological flyers and jumpers, which utilize intermittent actuation to rest muscles and maximize energy efficiency.

To overcome these limitations, we propose a Dynamics-Informed RL framework shown in Fig. 2 that bridges hybrid system modeling and data-driven control. We reframe the locomotion task from naive kinematic trajectory tracking to the stabilization of a hybrid limit cycle.

By embedding an Energy Manifold derived from the SLIP model into the reward structure via a mass-normalized Specific Energy, we ensure physically grounded optimization. By rewarding the phase-consistent behavior, it can make a bio-inspired impulse during stance. Furthermore, we integrate a high-fidelity, speed-dependent electromechanical actuator model into the training loop to induce the motors generate an efficient impulse. 

The main contributions of this work are summarized as follows:
\begin{itemize}
\item \textbf{Energy Reward Shaping:} We design a novel reward function grounded in active energy regulation. By targeting a nominal Specific Energy, we restrict optimization to a physically viable energy manifold, preventing ``hover-hopping'' and inducing a stable periodic limit cycle across continuous flight and discrete stance phases.
\item \textbf{Phase Consistent Reward Shaping:} Damping loss is counteracted by injecting net positive work into the system. A simplified model is assumed, in which the particle moves strictly vertically, and the energy balance principle is required, ultimately achieving phase consistency naturally without the heuristic state machine.
\item \textbf{Efficient Actuation Reward Shaping:} By integrating a rigorous nonlinear electrical motor model into the RL formulation and explicitly penalizing true electrical power losses, the agent naturally abandons inefficient actuation in favor of energy-efficient burst energy.
\end{itemize}

\begin{figure*}[htbp]
    \centering
    \includegraphics[width=0.9\linewidth]{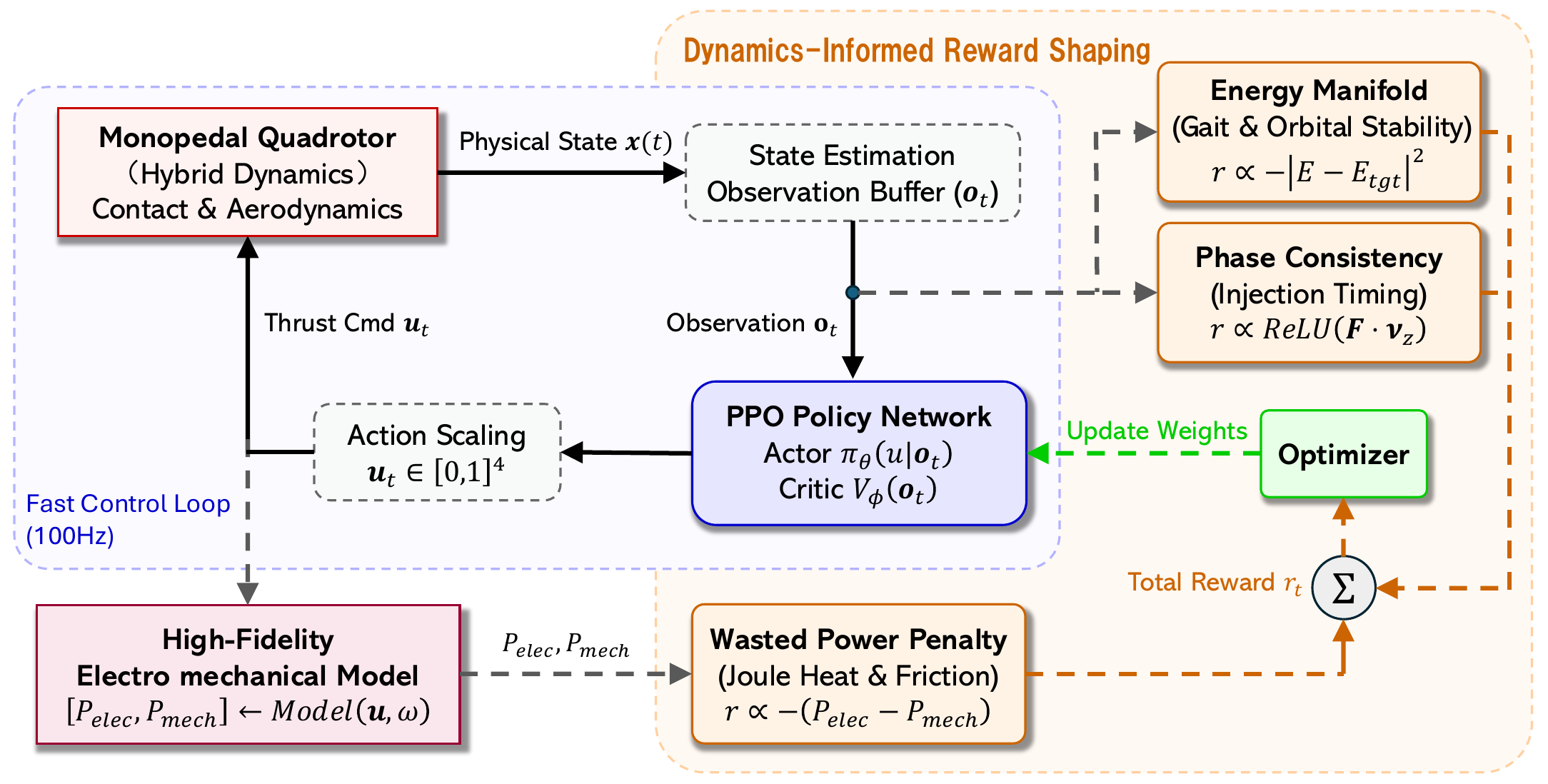} 
    \caption{\textbf{The proposed Dynamics-Informed Reinforcement Learning framework.} The system features a fast control loop (light violet region) interacting with the hybrid hopping environment. The core innovation lies in the Dynamics-Informed Reward Shaping module (orange region), which structurally informs the learning process. Concurrently, to ensure true hardware-level efficiency, actions are passed through a high-fidelity electro-mechanical actuator model to explicitly penalize the dissipated ``wasted power'' ($P_{waste} = P_{elec} - P_{mech}$).}
    \label{fig:system_overview}
\end{figure*}

% =================================================================================
% SECTION II: METHODOLOGY
% =================================================================================
\section{METHODOLOGY}

In this section, we formulate the agile and energy-efficient hopping control problem. We first introduce the hybrid dynamics and the high-fidelity actuator physics. We then reframe the locomotion task not as trajectory tracking, but as a Total Mechanical Energy Regulation problem, which is solved using Proximal Policy Optimization (PPO).

% ---------------------------------------------------------------------------------
% A. Hybrid Dynamics and Parameter Sensitivity
% ---------------------------------------------------------------------------------
\subsection{Hybrid Dynamics and Parameter Sensitivity}
The system operates on a hybrid automaton defined by the tuple $\mathcal{H}_{sys} = (\mathcal{D}, \mathcal{U}, \Delta)$. $\mathcal{D}$ is the domain of continuous states, $\mathcal{U}$ is the set of admissible controls, and $\Delta$ is the discrete reset map. The state space is partitioned into flight and stance phases by the guard function $\phi(\mathbf{q})=0$, representing the foot height derived from the generalized coordinate vector $\mathbf{q} = [p_x, p_y, p_z, \theta_x, \theta_y, \theta_z, l_s]^T$, where $p_x, p_y, p_z$ are position of the robot COM, and $\theta_x, \theta_y, \theta_z$ are the orientation of the robot using the intrinsic Z-Y-X sequence\cite{siciliano2009robotics}. $l_s$ is leg length, defined as the distance between the COM and the foot point, which can contact with ground. As defined in Fig. 1B, $\mathbf{q}$ is expressed in the world coordinate system. This explicitly aligns our analytical formulation with the measurement space of standard motion capture systems, paving the way for seamless sim-to-real physical deployment.

A discrete state transition occurs at touchdown ($\phi(\mathbf{q})=0$ and $\dot\phi(\mathbf{q})<0$). To accurately capture the collision physics without erroneously implying tangential elasticity, we must decompose the contact restitution. The post-impact velocity $\dot{\mathbf{q}}^+$ is related to the pre-impact velocity $\dot{\mathbf{q}}^-$ via the Saltation Matrix $\Delta$:
\begin{equation}
    \dot{\mathbf{q}}^+ = \Delta(\mathbf{q}, \bm{\xi}) \dot{\mathbf{q}}^- = \left( \mathbf{I} - \mathbf{M}^{-1}\mathbf{J}_c^T \bm{\Lambda}_{eff} (\mathbf{I} + \mathbf{E}) \mathbf{J}_c \right) \dot{\mathbf{q}}^-
\end{equation}
where $\mathbf{I}$ is the identity matrix, $\mathbf{M}$ is the generalized mass matrix, and $\mathbf{J}_c \in \mathbb{R}^{3 \times 7}$ is the full contact Jacobian. $\bm{\Lambda}_{eff} = (\mathbf{J}_c \mathbf{M}^{-1} \mathbf{J}_c^T)^{-1}$ is the operational space inertia matrix at the contact point. Crucially, $\mathbf{E} = \text{diag}(e, 0, 0)$ is the restitution matrix, which strictly applies the coefficient of restitution $e$ to the normal direction while enforcing a perfectly inelastic no-slip condition in the tangential directions. The vector $\bm{\xi} = [e, m]^T$ encapsulates uncertain environmental parameters: the coefficient of restitution $e$ and the robot mass $m$.

A critical challenge is the ill-conditioning of the sensitivity Jacobian $\mathbf{J}_{\xi} = \partial \dot{\mathbf{q}}^+ / \partial \bm{\xi}$. As established in non-smooth mechanics \cite{brogliatononsmooth}, for stiff robotic legs, the impact mapping becomes highly sensitive to parameter variations. Consequently, infinitesimal estimation errors in $\bm{\xi}$ lead to large divergences in the post-impact trajectory. This mathematical property highlights the fragility of explicit model-based planning under parametric uncertainty. Therefore, Eq. (1) formally motivates our design choice: rather than attempting to explicitly invert this fragile impact map, our model-free RL framework, guided by dynamics-informed reward, organically develops policies that are inherently robust to such ill-conditioned impact dynamics.

% ---------------------------------------------------------------------------------
% C. Actuator Dynamics and Efficiency Modeling
% ---------------------------------------------------------------------------------
\subsection{Actuator Dynamics and Efficiency Modeling}
To rigorously optimize energy efficiency, we model the specific hardware topology consisting of coreless DC Motors driven by MOSFETs. In our framework, the policy outputs a normalized thrust command (throttle) $u \in [0, 1]$, which maps linearly to the rotor's physical thrust force: $F_{thrust} = u \cdot F_{max}$, where $F_{max}$ is the maximum thrust force of a single rotor.

\subsubsection{High-Fidelity Electromechanical Model}
The rotor angular velocity is derived from the aerodynamic mapping $\omega = \sqrt{F_{thrust} / K_f}$, where $K_f$ is the thrust coefficient. Derived from the standard DC motor torque balance equation \cite{krause2002analysis}, the average armature current $I_a$ is coupled with the aerodynamic drag torque $\tau_{load} = K_m \omega^2$ and the internal speed-dependent losses:
\begin{equation}
    I_a = \frac{\tau_{load}}{K_t} + I_{static} + D_v \omega
\end{equation}
where $K_m$ is the drag coefficient, $K_t$ is the torque constant, $I_{static}$ accounts for the static no-load current, and $D_v$ represents the dynamic friction and iron loss coefficient combined.

\subsubsection{Nonlinear Efficiency Function}
The total electrical power consumption $P_{elec}$ encompasses the Joule heating, the mechanical aerodynamic output, and the internal electromechanical losses:
\begin{equation}
    P_{elec} = \underbrace{I_a^2 R_{total}}_{\text{Joule Loss}} + \underbrace{\tau_{load} \omega}_{\text{Mechanical Output}} + \underbrace{(I_{static} + D_v \omega) \omega K_t}_{\text{Internal Friction \& Iron Loss}}
\end{equation}
Crucially, $R_{total}$ explicitly sums the motor's dynamic armature resistance ($R_{mtr} \approx 0.45 \Omega$) and the MOSFET on-resistance ($R_{ds(on)} \approx 0.085 \Omega$), reflecting the specific hardware parameters used in our platform introduced in Sec. III-A. 

This high-fidelity, physics-based model mathematically captures the nonlinear efficiency degradation at extreme high-speed regimes, naturally forming bounded regions of optimal efficiency. It reveals that the system efficiency $\eta = P_{mech}/P_{elec}$ is a nonlinear function of torque and speed. Minimizing $P_{elec}$ encourages the agent to explicitly exploit highly efficient, bounded operating regions, avoiding the severe efficiency drop caused by quadratic aerodynamic drag and internal friction.

% ---------------------------------------------------------------------------------
% D. Orbital Stabilization as Optimization
% ---------------------------------------------------------------------------------
\subsection{Orbital Stabilization as Optimization}
We reframe the stable hopping task as an energy-based optimization problem. The objective is to find a control law $\mathbf{u}(t) \in [0,1]^4$, representing the rotor thrust commands acting on the continuous flight dynamics, that stabilizes a limit cycle while minimizing energy cost.

\subsubsection{Objective I: Total Mechanical Energy Regulation}
Standard Euclidean tracking ($J = ||z - z_{tgt}||$) is ill-posed for hopping as it encourages hovering (zero velocity). Instead, we utilize the total energy $\mathcal{H}(\mathbf{x}) = mgz + \frac{1}{2}m v_z^2$, representing the vertical mechanical energy, where $\mathbf{x} = [z, v_z]^T$ represents the vertical state. 

To ensure the reward function remains scale-invariant and robust against inherent parametric uncertainties regarding the robot's mass $m$, we normalize the vertical mechanical energy by gravitational force to define the Specific Energy $E(\mathbf{x})$:
\begin{equation}
    E(\mathbf{x}) = \frac{\mathcal{H}(\mathbf{x})}{mg} = z + \frac{v_z^2}{2g}
\end{equation}

We define the target energy manifold $\mathcal{M}^*$ as the level set where $E(\mathbf{x}) = z_{tgt}$. The optimization objective is to minimize the squared energy error $V_{\mathcal{M}}$:
\begin{equation}
    \min_{\mathbf{u}} \int_{0}^{T} V_{\mathcal{M}}(\mathbf{x}) dt, \quad V_{\mathcal{M}} = \left(E(\mathbf{x}) - z_{tgt} \right)^2
\end{equation}
Where $T$ is the hopping period. In our RL implementation, this continuous integral is evaluated at discrete control steps. Minimizing this residual forces the state $\mathbf{x}$ to converge to the parabolic energy orbit. This naturally induces convergence to a stable periodic orbit with the desired peak height while preventing reward explosion or vanishing when training across varying mass distributions.

\subsubsection{Objective II: Phase Stability Condition}
To sustain a limit cycle, the system must inject net positive work to counteract damping. Assuming a simplified model where a point mass moves strictly vertically with a linear spring-damper, the Energy Balance Principle requires:
\begin{equation}
    \oint F_{thrust,z} \cdot v_z \, dt = \oint b v_z^2 \, dt
\end{equation}
where $b$ is the damping coefficient and $F_{thrust,z}$ is the vertical component of the thrust. This implies that to compensate for impact losses, the system must predominantly inject energy during the upward restitution phase ($v_z > 0$). We formulate this as a work maximization objective subject to unilateral constraints ($u \ge 0$). The objective is to maximize the Positive Power Flow $\Psi$:
\begin{equation}
    \max_{\mathbf{u}} \int_{0}^{T} \Psi(\mathbf{u}, v_z) dt, \quad \Psi = \text{ReLU}( F_{thrust,z}(\mathbf{u}) \cdot v_z )
\end{equation}
where $\text{ReLU}(x) = \max(0, x)$ is the Rectified Linear Unit. This objective ensures that actuation occurs strictly when the thrust aligns with the velocity (\textit{i.e.}, the restitution phase, $v_z > 0$), naturally emerging phase consistency without heuristic state machines.

% ---------------------------------------------------------------------------------
% E. Robust Learning via PPO
% ---------------------------------------------------------------------------------
\subsection{Robust Learning Implementation}
The combined optimization problem is non-convex and involves severe contact discontinuities.

\subsubsection{PPO Concept}
We employ PPO, maximizing a clipped surrogate objective:
\begin{equation}
    L^{CLIP}(\theta) = \hat{\mathbb{E}}_t \left[ \min(r_t(\theta) \hat{A}_t, \text{clip}(r_t(\theta), 1-\epsilon, 1+\epsilon)\hat{A}_t) \right]
\end{equation}
\textbf{Advantage for Hybrid Systems:} In hopping locomotion, discrete impact events cause destructive discontinuities in the state-value function $V(s)$. Crucially, PPO's clipping mechanism structurally bounds the policy update step size, ensuring gradient stability despite the high sensitivity of hybrid contact dynamics.

\subsubsection{Mapping Optimization to Rewards}
The theoretical objectives derived in Sec. II-C directly map to the cumulative reward $R = r_{manifold} + r_{phase} + r_{eff}$:

\begin{itemize}
    \item \textbf{Energy Manifold Reward ($r_{manifold} \leftrightarrow$ Obj. I):}
    We employ a Gaussian kernel to cast energy tracking as a maximization problem, actively stabilizing the system onto the target energy manifold $\mathcal{M}^*$:
    \begin{equation}
        r_{manifold} = w_e \exp\left( -\lambda_e \left( E(\mathbf{x}) - z_{tgt} \right)^2 \right)
    \end{equation}

    \item \textbf{Phase Consistent Reward ($r_{phase} \leftrightarrow$ Obj. II):}
    To encourage energy injection strictly during stance, we reward positive mechanical power:
    \begin{equation}
        r_{phase} = w_p \cdot \text{ReLU}(\bar{u} v_z)
    \end{equation}
    where $\bar{u}$ proxies the normalized vertical thrust. Rather than relying on rigid state machines to prevent ``reward hacking,'' energy injection is organically bounded: thrusting against the velocity ($v_z < 0$) yields negative mechanical work, which inherently amplifies the subsequent efficiency penalty ($r_{eff}$).

    \item \textbf{Wasted Power Penalty ($r_{eff} \leftrightarrow$ Sec. II-B):}
    Instead of naively penalizing total control effort, we selectively penalize non-conservative wasted energy based on our high-fidelity actuator model:
    \begin{equation}
        r_{eff} = -w_{cost} \sum_{i=1}^{4} \left( P_{elec}^{(i)} - P_{mech}^{(i)} \right)
    \end{equation}
    where $P_{elec}^{(i)}$ and $P_{mech}^{(i)}$ are the electrical input and mechanical output power, respectively. By explicitly punishing pure energy waste (e.g., Joule heating), this formulation forces the agent to exploit highly efficient motor regimes and natively abandon continuous, inefficient aerial actuation.
\end{itemize}

\subsubsection{Observation Space}
To enable the solver to compute these objectives, the policy input vector $\mathbf{o}_t \in \mathbb{R}^{18}$ fed to the PPO network is formulated by augmenting the robot's physical state with the commanded goals:
\begin{equation}
    \mathbf{o}_t = [\mathbf{s}_t, \mathbf{g}_t]^T
\end{equation}
where the intrinsic robot state is $\mathbf{s}_t = [ z, \mathbf{q}_{quat}, \mathbf{v}_{\mathcal{B}}, \bm{\omega}_{\mathcal{B}}, \mathbf{u}_{t-1}, I_{contact} ]^T$, and the external goal vector is $\mathbf{g}_t = [v_{x}^{cmd}, z_{tgt}]^T$. Here, $z$ is the current height, $\mathbf{q}_{quat}$ is the body orientation in quaternions, $\mathbf{v}_{\mathcal{B}}$ and $\bm{\omega}_{\mathcal{B}}$ are the linear and angular velocities in the body frame, $\mathbf{u}_{t-1}$ is the previous action, and $I_{contact}$ is the contact indicator. The target height $z_{tgt}$ and commanded forward speed $v_{x}^{cmd}$ are strictly required as conditions for the policy to satisfy the specific energy and kinematic tracking rewards.

% =================================================================================
% SECTION III: EXPERIMENTS AND RESULTS
% =================================================================================

\begin{table}[h]
\centering
\caption{Physical and High-Fidelity Actuator Parameters of Hopping Quadcopter}
\label{tab:params}
\begin{tabular}{lcc}
\hline
\textbf{Parameter} & \textbf{Symbol} & \textbf{Value} \\ 
\hline
\multicolumn{3}{c}{\textbf{Robot Body \& Leg Dynamics}} \\
\hline
Total Mass & $m$ & 0.099 kg \\
Leg Stiffness & $k_s$ & 75 N/m \\
Leg Damping & $b_s$ & 0.05 Ns/m \\
Leg Rest Length & $l_0$ & 0.12 m \\
Control Frequency & $f_{ctrl}$ & 100 Hz ($\Delta t = 0.01$ s) \\
\hline
\multicolumn{3}{c}{\textbf{8520 Coreless Motor \& Propeller Model}} \\
\hline
Battery Voltage & $V_{bat}$ & 3.7 V \\
Dynamic Motor Resistance & $R_{mtr}$ & 0.45 $\Omega$ \\
MOSFET On-Resistance & $R_{ds(on)}$ & 0.085 $\Omega$ \\
Motor Time Constant & $\tau$ & 0.023 s \\
Motor Velocity Constant & $K_v$ & 13500 RPM/V \\
Motor Torque Constant & $K_t$ & $7.07 \times 10^{-4}$ N$\cdot$m/A \\
Thrust Coefficient & $K_f$ & $3.27 \times 10^{-8}$ N/(rad/s)$^2$ \\
Drag Torque Coefficient & $K_m$ & $1.42 \times 10^{-10}$ N$\cdot$m/(rad/s)$^2$ \\
Max Angular Velocity & $\omega_{max}$ & 3665.0 rad/s \\
Static No-Load Current & $I_{static}$ & 0.05 A \\
Dynamic Loss Coefficient & $D_v$ & $4.09 \times 10^{-5}$ A/(rad/s) \\
\hline
\end{tabular}
\end{table}

\begin{table}[h]
\centering
\caption{Hyperparameters for PPO Training and Reward Formulation}
\label{tab:hyperparams}
\resizebox{\columnwidth}{!}{
\begin{tabular}{@{}llc@{}}
\toprule
\textbf{Category} & \textbf{Parameter} & \textbf{Value} \\ 
\midrule
\multirow{12}{*}{\textbf{PPO}} 
& Algorithm & PPO (SB3) \\
& Architecture & MLP [128, 128], Tanh \\
& Learning Rate & $1 \times 10^{-4}$ \\
& Buffer Size & 4096 \\
& Batch Size & 512 \\
& Epochs & 10 \\
& Discount Factor ($\gamma$) & 0.99 \\
& GAE Lambda ($\lambda$) & 0.95 \\
& Clip Range ($\epsilon$)& 0.2 \\
& Entropy Coef. & 0.01 \\
& Total Timesteps & $3 \times 10^7$ \\ 
& Parallel Envs & 8 \\ 
\midrule
\multirow{6}{*}{\textbf{Reward}} 
& Manifold weight ($w_e$) & 6.0 \\
& Manifold scale ($\lambda_e$) & 10.0 \\
& Power inject weight ($w_p$) & 5.0 \\
& Efficiency penalty ($w_{cost}$) & 0.1 \\
\bottomrule
\end{tabular}%
}
\end{table}

\begin{figure}[htbp]
    \centering
    \includegraphics[width=1\linewidth]{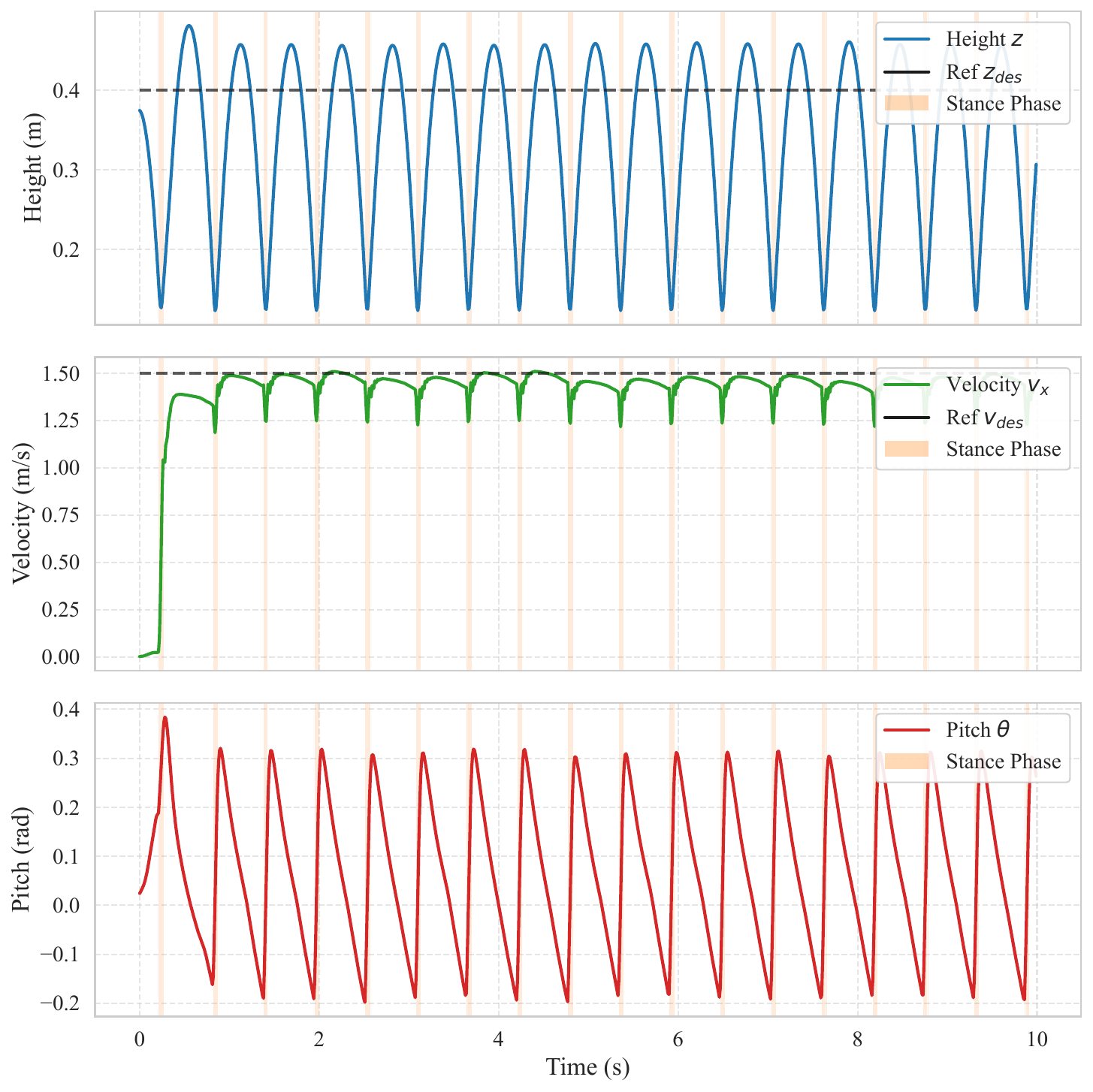}
    \caption{\textbf{Kinematic response of the hopping quadcopter tracking a target velocity of $1.5$\,m/s.} The shaded regions represent the stance phase (contact). The agent successfully modulates body pitch to regulate forward speed while maintaining a stable hopping height.}
    \label{fig:kinematics}
\end{figure}

\section{Simulation Results}
\label{sec:experiments}

To validate the proposed Dynamics-Informed RL framework, we conducted extensive simulations in the MuJoCo physics engine. The simulation frequency was set to 1000 Hz, while the control policy operated at 100 Hz. We utilized the PPO algorithm implemented in \textit{Stable-Baselines3}.

\subsection{Simulation Setup}
The simulated robot model is based on the Crazyflie 2.1 quadcopter available in the laboratory, modified with a custom 3D-printed compliant leg. The propulsion system consists of four coreless DC motors (model: 8520) driven by MOSFETs (model: SI2302), powered by a single-cell 3.7 V LiPo battery. The total system mass is 99 g.
Key physical parameters derived from our model and system identification are listed in Table~\ref{tab:params}. The leg damping coefficient ($b_s=0.05$ Ns/m) was identified to be significant, posing a challenge for energy maintenance. The RL policy network consists of a Multi-Layer Perceptron (MLP) with two hidden layers of 128 units each, using Tanh activation functions. The detailed hyperparameters and reward formulation used for training are summarized in Table~\ref{tab:hyperparams}.

\begin{figure}[htbp]
    \centering
    \includegraphics[width=1\linewidth]{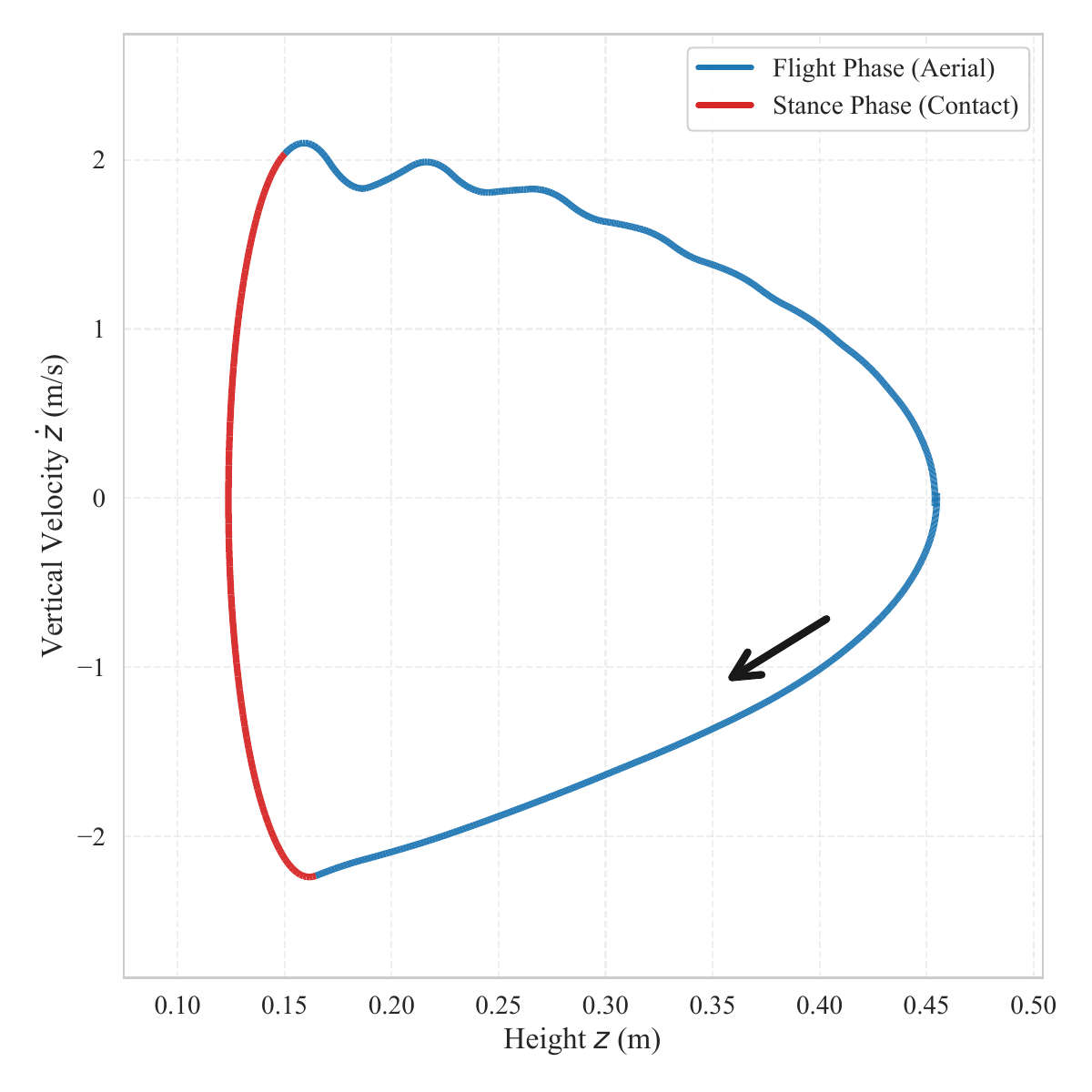}
    \caption{\textbf{Vertical phase portrait ($z$ vs. $\dot{z}$).} The system converges to a stable, closed limit cycle, indicating a perfect balance between energy injection (thrust) and dissipation (damping/impact).}
    \label{fig:limit_cycle}
\end{figure}

\subsection{Agile Monopedal Hopping Performance}
We first evaluate the fundamental locomotion capability. The robot was commanded to track a forward velocity of $1.5$ m/s while maintaining a target apex height of $0.4$ m.

\textbf{Kinematic Tracking:} 
As shown in Fig.~\ref{fig:kinematics}, the robot exhibits stable periodic hopping. The vertical height (Top) strictly follows the flight-stance phases, and the forward velocity (Middle) converge to the reference value within 2 seconds. Although the emergent apex height exhibits a slight, constant offset above the nominal target $z_{tgt} = 0.4$\,m, this phenomenon highlights a core advantage of our framework: rather than rigidly enforcing a kinematic spatial boundary---which would require inefficient, high-frequency control efforts---the policy organically converges to a globally optimal energy manifold. Notably, the pitch angle (Bottom) shows a clean, periodic oscillation synchronized with the hopping cycle, which is a behavior that emerges naturally from our velocity tracking reward combined with the energy manifold.

\textbf{Limit Cycle Stabilization:}
To verify the orbital stability, we analyze the phase portrait of the vertical states ($z, \dot{z}$). Fig.~\ref{fig:limit_cycle} visualizes the system's trajectory in phase space over 10 seconds. The convergence to a single, closed loop indicates that the Dynamics-Informed reward successfully guided the agent to find a stable periodic orbit. The trajectory is smooth and repeatable, demonstrating robustness against the hybrid discrete transitions at impact.

\subsection{Energy Efficiency and Mechanism Analysis}
\begin{figure}[htbp]
    \centering
    \includegraphics[width=1\linewidth]{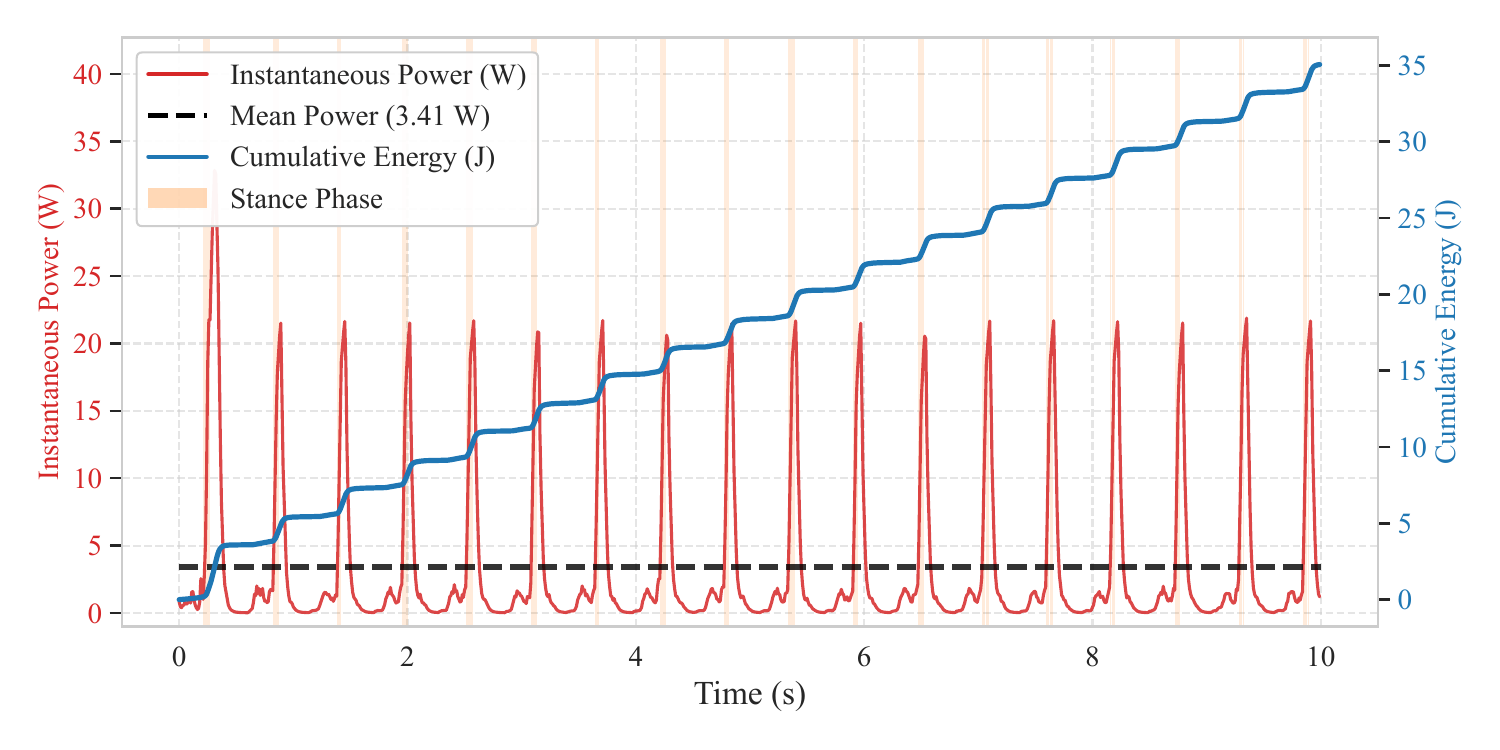}
    \caption{\textbf{Energy efficiency analysis.} The power profile (red) reveals a pulsed actuation strategy: the agent minimizes thrust during flight and injects energy explosively during the stance restitution phase, mimicking biological hopping.}
    \label{fig:hopping_power}
\end{figure}

We then analyze whether the learned policy aligns with our theoretical hypothesis regarding energy injection. Fig.~\ref{fig:hopping_power} presents the instantaneous power, mean power, and cumulative energy. The mean power reported is calculated exclusively over the steady-state limit cycle, deliberately truncating the initial $30\%$ of the transient data. A distinct ``pulse'' actuation pattern is observed: power consumption is minimal during the ballistic flight phase and peaks sharply only during the restitution phase of the stance. This confirms that the agent organically utilizes an ``Actuation Burst'' strategy. By ``coasting'' in the air and ``pushing'' strictly when the spring is extending, the system maximizes the positive mechanical work done by the motors without requiring heuristic state machines or artificial action truncation commonly relied upon in previous hopping systems \cite{poulakakis2009spring, haldane2017repetitive, zhu2022pogodrone, bai2024agile}.

\begin{figure}[htbp]
    \centering
    \includegraphics[width=1\linewidth]{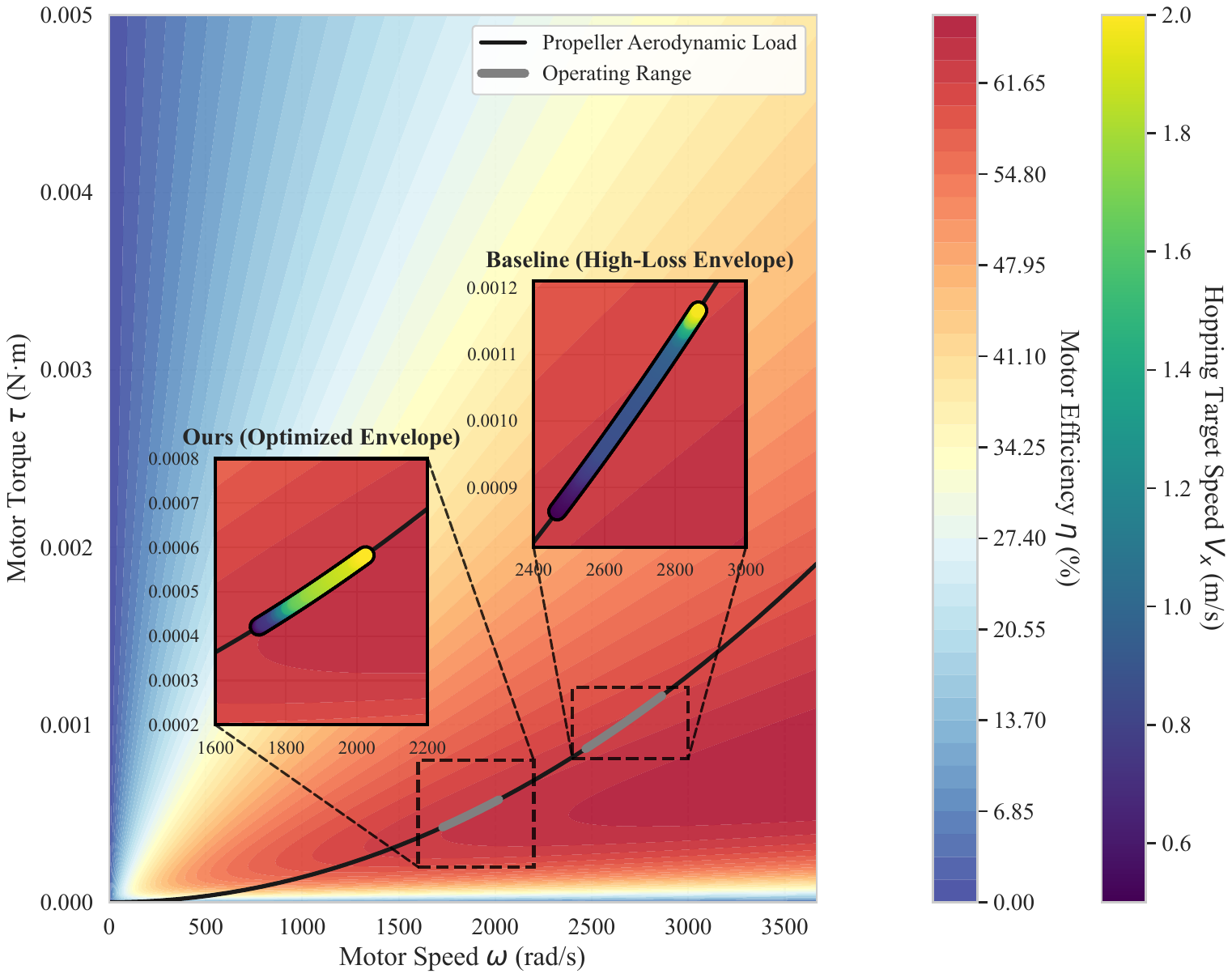} 
    \caption{\textbf{Operating Envelope Comparison on the Actuator Efficiency Map.} The color-graded ribbons represent the equivalent operating points during the active stance phase across varying hopping speeds ($V_x \in [0.5, 2.0]$ m/s). \textbf{Right Inset:} Without the efficiency penalty, the baseline blindly operates in a high-loss, high-speed regime. \textbf{Left Inset:} Our proposed dynamics-informed RL actively shifts the operating envelope into the optimal high-efficiency sweet spot, demonstrating profound system-level energy awareness.}
    \label{fig:efficiency_map}
\end{figure}

\subsection{Actuator Operating Envelope and Efficiency}
To explicitly validate the efficacy of penalizing the wasted power derived from our detailed electromechanical model, we conduct a comparative analysis of the motor's operating states during the critical stance phase. We evaluate the proposed policy against a standard baseline policy trained without the wasted-power penalty across varying target velocities ($0.5$ to $2.0$ m/s). As shown in the Fig.~\ref{fig:efficiency_map} (calculated using the nonlinear function $\eta = P_{mech}/P_{elec}$ derived in Sec. II-B), the operating trajectories of both agents strictly adhere to the 1D aerodynamic propeller load curve ($\tau \propto \omega^2$), yet their learned operating envelopes diverge significantly. 

\textbf{Baseline (High-Loss Envelope):} Driven without the efficient actuation rewards, the baseline agent (right inset) naively exploits the motor's maximum thrust capacity to minimize stance duration. It converges to a high-speed regime ($\approx 2400-3000$ rad/s). In this region, despite generating massive instantaneous impulses, the severe $I^2R$ copper losses and dynamic internal friction drastically degrade the overall energy efficiency, trapping the agent in a thermally prohibitive local optimum.

\textbf{Ours (Optimized Envelope):} Conversely, guided by the physics-informed wasted-power penalty ($P_{waste} = P_{elec} - P_{mech}$), our proposed framework (left inset) strategically shifts its operating envelope downwards. The agent organically discovers the motor's true efficiency sweet spot ($\approx 1600-2200$ rad/s). By accurately perceiving the nonlinear surge in Joule heating at high speeds, the policy learns to perfectly balance the aerodynamic thrust requirements with the electromechanical thermal limits. This compellingly proves that our framework successfully optimizes the true global limit cycle and hardware-software system efficiency, entirely bypassing the need for artificial action truncation or heuristic state machines commonly used in previous hopping controllers \cite{raibert1986legged, haldane2017repetitive, zhu2022pogodrone, csomay2023nonlinear}.

\subsection{Comparative Analysis: Stability and Robustness}
\begin{figure}[htbp]
    \centering
    \includegraphics[width=1\linewidth]{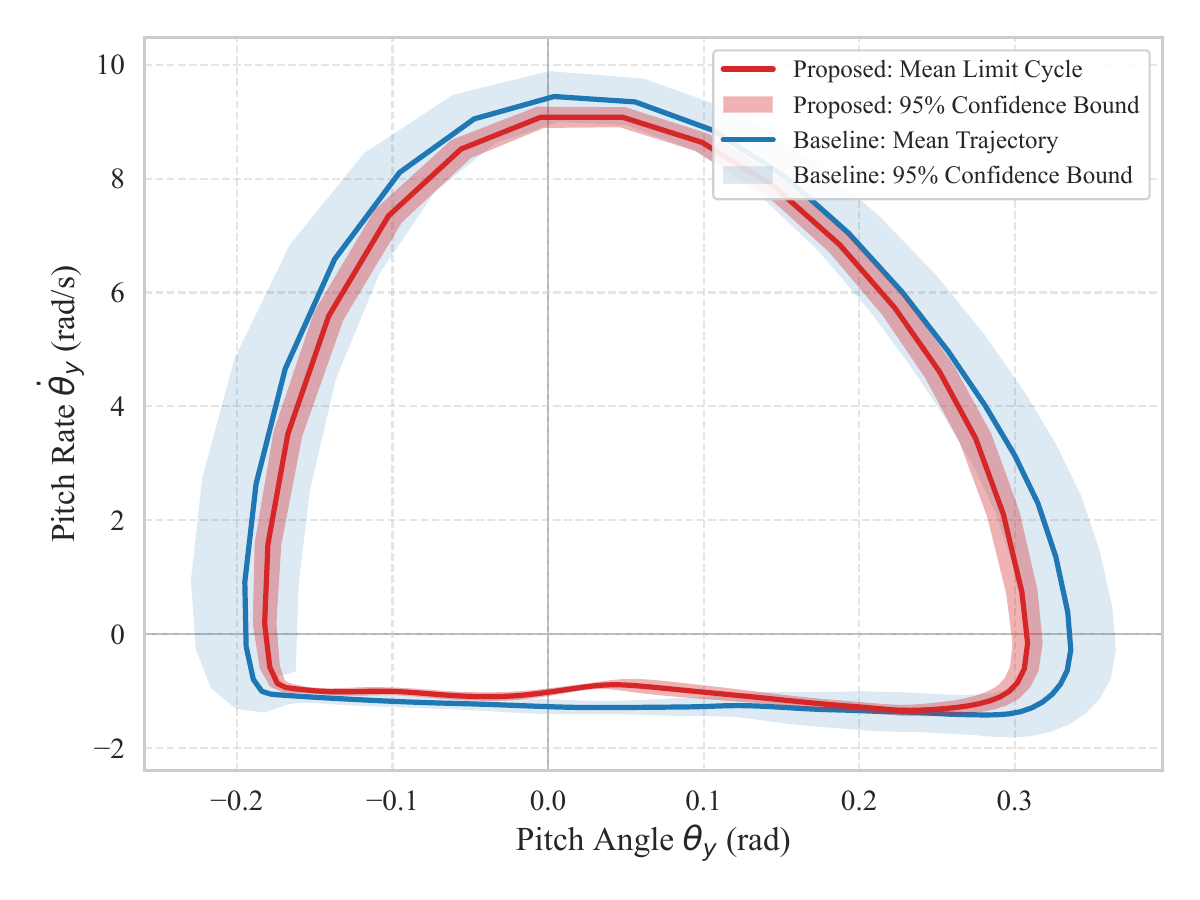}
    \caption{\textbf{Phase portrait of the pitch dynamics ($\theta_y$ vs. $\dot{\theta}_y$) during steady-state high-speed hopping.} Solid lines denote the mean limit cycles computed over multiple normalized consecutive jumps, while the shaded regions encompass the $95\%$ confidence bounds of the cycle-to-cycle state distribution. The proposed method (Red) achieves a highly compact variance envelope, demonstrating strict orbital stability compared to the widely dispersed baseline (Blue).}
    \label{fig:phase_portrait}
\end{figure}

To demonstrate the necessity of the proposed framework, we compare it against baselines lacking the phase-consistent reward. To rigorously evaluate the orbital stability of the hopping gaits, Fig.~\ref{fig:phase_portrait} illustrates the phase portrait of the pitch dynamics ($\theta_y$ vs. $\dot{\theta}_y$). By segmenting the continuous steady-state trajectories into individual normalized hopping cycles, we extract the mean limit cycles (solid lines) and their corresponding $95\%$ confidence bounds (shaded regions). As visually evident, the proposed phase-consistent method (Red) establishes an extremely tight confidence envelope around its mean limit cycle. This proves its superior resilience against compounding integration errors and its strict orbital stability over consecutive jumps. Conversely, the baseline method (Blue) exhibits a massively dispersed variance envelope. It is crucial to emphasize that both policies were trained to full convergence; thus, this cycle-to-cycle divergence exposes a fundamental structural limitation of standard kinematic rewards. Without the work maximization objective explicitly mapping energy injection to the spring's restitution phase, the baseline agent haphazardly outputs thrust, severely destabilizing the attitude dynamics and failing to maintain a repeatable limit cycle.

\textbf{Necessity of the Energy Manifold Reward:}
Next, we investigate the fundamental role of the Energy Manifold formulation. We trained a naive baseline where the dynamics-informed reward was replaced by a standard instantaneous position tracking penalty.
\begin{equation}
r_{naive} = w_e \exp\left( -\lambda_e |z - z_{tgt}| \right)
\end{equation}

\begin{figure}[htbp]
    \centering
    \includegraphics[width=1\linewidth]{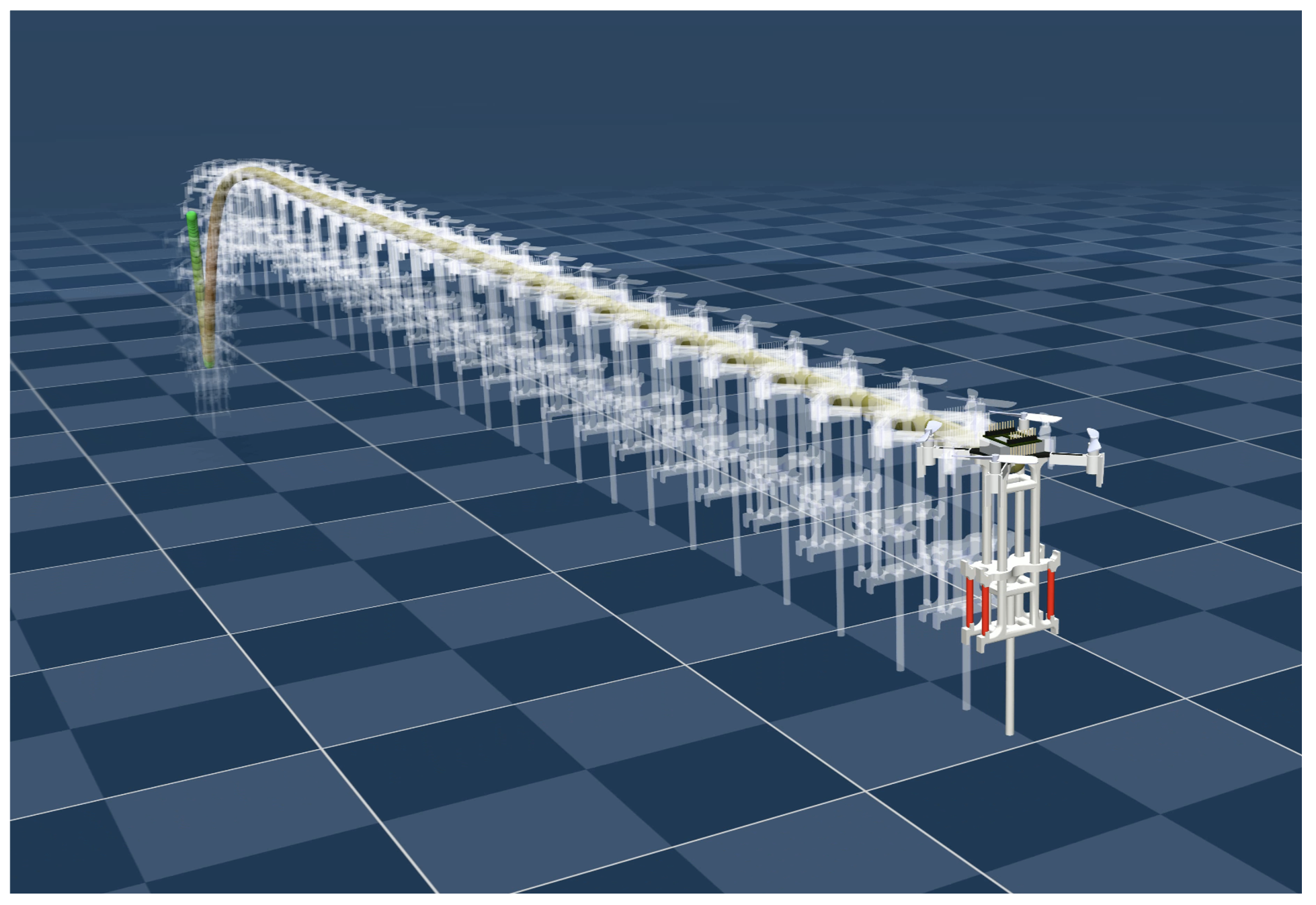} 
    \caption{\textbf{Simulation snapshot of the baseline agent.} The agent resorts to continuous aerial flight without ground interaction, verifying that standard spatial tracking rewards trigger severe reward hacking in hybrid systems.}
    \label{fig:ablation_energy_screenshot}
\end{figure}

The simulation snapshot shown in Fig.~\ref{fig:ablation_energy_screenshot}, which visually confirms this ``hover-hopping'' behaviour, and the leg remains extended without ground contact, reveals a critical failure mode driven by reward hacking \cite{yuan2019novel, ibrahim2024comprehensive, knox2024learning}.  Constrained by the strict position error penalty, the policy converges to a local optimum: it exploits the quadcopter's high thrust-to-weight ratio to hover statically at the target altitude ($z \approx z_{des}, \dot{z} \approx 0$) while moving forward.  In contrast, our Specific Energy formulation successfully induces a stable hopping cycle shown in Fig. 3, proving that decoupling energy regulation from instantaneous spatial tracking is essential for hybrid locomotion.

\subsection{Energy Advantage: Hopping vs. Hovering}

\begin{figure}[htbp]
    \centering
    \includegraphics[width=1\linewidth]{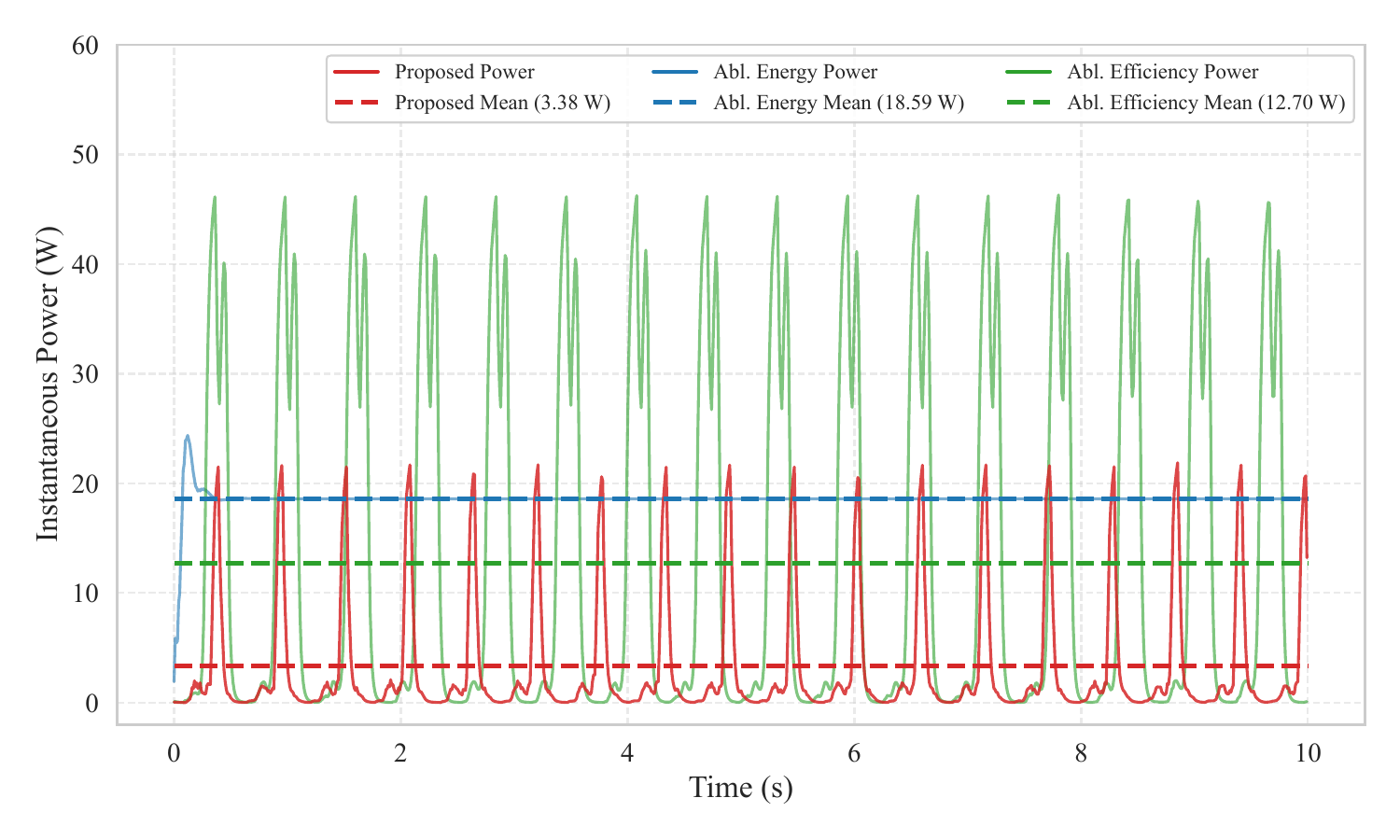}
    \caption{Quantitative power comparison during a 10-second steady-state hopping phase. Instantaneous power (solid lines) and the corresponding mean power (dashed lines). The red line represents the power under the Dynamics-Informed reward; the blue line represents the power without the energy manifold reward; and the green line represents the power without the efficiency penalty reward.}
\label{fig:power_comparison}
\end{figure}

To explicitly evaluate the energetic contribution of the manifold tracking objective and efficiency penalty, we compare the proposed policy against a baseline trained without the dynamics-informed guidance. As illustrated in Fig.~\ref{fig:power_comparison}, without the structural guidance of $r_{manifold}$, the baseline agent fundamentally fails to exploit the passive compliance of the spring-loaded leg. It relies on continuous flying to forcefully maintain locomotion, resulting in a high mean power of $\approx \mathbf{18.59}$\,W. Without the guidance of $r_{eff}$, the baseline agent's motor blindly operates in a high-loss, high-speed regime, resulting in a high mean power of $\approx \mathbf{12.7}$\,W.

In sharp contrast, the proposed policy, driven directly by $r_{manifold}$, autonomously converges to an energy-optimal ``pulsed actuation'' strategy. By forcing the agent to track the nominal energy manifold, it learns to restrict major energy injection strictly to the brief stance restitution phase and perfectly leverages ballistic aerial coasting. This physically synergistic behaviour drastically reduces the mean power to $\approx \mathbf{3.32}$\,W. The proposed policy can reduce up to $\mathbf{82}\%$ and $\mathbf{73}\%$ power consumption compared with that without the $r_{manifold}$ and $r_{eff}$ reward. This rigorously demonstrates that the Dynamics-Informed reward is the fundamental driver of energetic superiority, seamlessly orchestrating the passive dynamics to minimize the total energy expenditure.

\begin{figure}[htbp]
    \centering
    \includegraphics[width=1\linewidth]{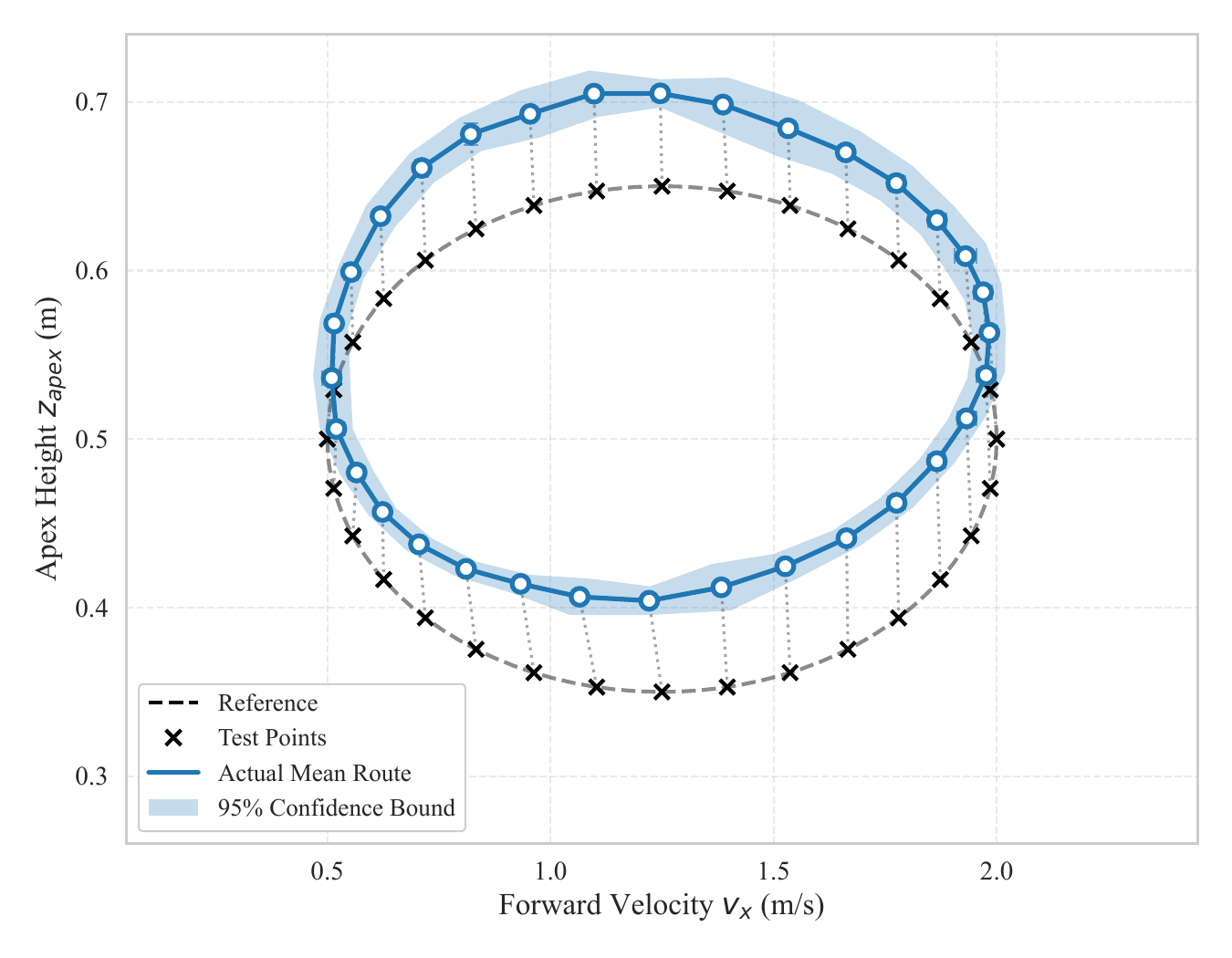}
    \caption{Statistical stability analysis across varying target velocities ($N=10$ trials). The agent maintains safe and consistent height regulation at different velocities with a bounded steady-state offset. The policy demonstrates exceptional velocity tracking stability and accuracy across the entire tested speed range ($0.5$-$2.0$\,m/s)}.
    \label{fig:stability}
\end{figure}

\subsection{Stability and Energy-Kinematics Trade-off}

We evaluate the policy's generalization across target speeds ($0.5-2.0\ m/s$) and apex heights ($0.35-0.65\ m$). As shown in Fig.~\ref{fig:stability}, the agent achieves continuous velocity tracking up to $2.0\ m/s$, rapidly converging to a robust periodic limit cycle. While recent controllers can execute high-speed transient leaps \cite{li2025high}, maintaining such speeds over extended trajectories remains an open challenge. Sustained tracking is typically constrained to low speeds ($\sim0.2\ m/s$) to prevent accumulative errors from severe attitude-contact coupling \cite{zhu2022pogodrone, bai2024agile, li2025high}. Conversely, our end-to-end Energy Manifold organically orchestrates the hybrid dynamics. This ensures the asymptotic stability of the high-speed limit cycle and averts the catastrophic height degradation common in sustained agile hopping \cite{huang2024real}

Interestingly, an emergent steady-state offset ($\sim0.05\ m$, or $12.5\%$) below the target apex height is consistently observed. Rather than a control failure, this reflects a physics-informed trade-off. At higher speeds, the stance phase drastically shortens. Eliminating this minor positional error would require massive peak thrusts during the brief contact window. Because internal electrical losses scale quadratically with current ($I^2R$), such impulses are strictly penalized by our efficiency reward. Consequently, the policy organically sacrifices marginal height accuracy to constrain the motors within their optimal efficiency envelope.

\section{Conclusion}

We present a Dynamics-Informed Deep Reinforcement Learning framework for a monopedal hopping quadcopter. By tracking a target Specific Energy manifold and penalizing true electro-mechanical waste, our approach eradicates the ``reward hacking'' typical of unconstrained RL. The agent organically learns a bio-inspired pulsed actuation strategy, injecting energy strictly during spring restitution without heuristic state machines. Simulations validate robust velocity tracking up to $2.0 m/s$ and demonstrate staggering energy reductions of $82\%$ and $73\%$ compared to continuous hovering and naive RL baselines. This proves that embedding physical laws into neural network optimization unlocks highly efficient hybrid locomotion.

\addtolength{\textheight}{-12cm}   % This command serves to balance the column lengths

%%%%%%%%%%%%%%%%%%%%%%%%%%%%%%%%%%%%%%%%%%%%%%%%%%%%%%%%%%%%%%%%%%%%%%%%%%%%%%%%

%%%%%%%%%%%%%%%%%%%%%%%%%%%%%%%%%%%%%%%%%%%%%%%%%%%%%%%%%%%%%%%%%%%%%%%%%%%%%%%%

\bibliographystyle{IEEEtran}
\bibliography{IEEEexample}

\end{document}